# ADAPTIVE ENSEMBLE LEARNING WITH GAUSSIAN COPULA FOR LOAD FORECASTING

*Junying Yang[1], Gang Lu[2*], Xiaoqing Yan[3], Peng Xia[4], Di Wu[5]*

[1]*College of Computer and Information Science, Southwest University, Chongqing 400715, China. yjy123@email.swu.edu.cn.*

[2]*Energy Strategy and Planning Research Department, State Grid Energy Research Institute Co., Ltd., Beijing 102209, China. lugang@sgeri.sgcc.com.cn.*

[3]*Energy Strategy and Planning Research Department, State Grid Energy Research Institute Co., Ltd., Beijing 102209, China. yanxiaoqing@sgeri.sgcc.com.cn.*

[4]*Energy Strategy and Planning Research Department, State Grid Energy Research Institute Co., Ltd., Beijing 102209, China. xiapeng@sgeri.sgcc.com.cn.*

[5] *College of Computer and Information Science, Southwest University, Chongqing 400715, China. wudi1986@swu.edu.cn.*

\* Corresponding Author

## Abstract

Machine learning (ML) is capable of accurate Load Forecasting from complete data. However, there are many uncertainties that affect data collection, leading to sparsity. This article proposed a model called Adaptive Ensemble Learning with Gaussian Copula to deal with sparsity, which contains three modules: data complementation, ML construction, and adaptive ensemble. First, it applies Gaussian Copula to eliminate sparsity. Then, we utilise five ML models to make predictions individually. Finally, it employs adaptive ensemble to get final weighted-sum result. Experiments have demonstrated that our model are robust.

## 1 Introduction

With the rapid development of the power system, the prediction of power consumption has become increasingly significant. Therefore, the realization of rational planning and efficient utilization of energy has become an important issue that needs to be solved at present urgently. LF uses scientific modelling methods to fully explore historical features and predict future. However, many uncertainties, such as tariff policy, will lead to the stochastic nature of LF[1]. Besides, with the rapid development of ultra-high voltages and new energies, not only multiple power sources are generated, but also the LF becomes less predictable[2]. Fortunately, ML developed recently shows the advantage of personalization of fine-grained slots. For example, a series of decision trees are connected to obtain a deeply structured model after extracting the deep latent factors of the high-dimensional and sparse matrices[3]. Besides, a novel outlier-resilient autoencoder (ORA) derived by deep neural network (DNN) can removing anomaly and effectively against sparsity[4]. However, the natural characteristics of incomplete data collection due to sensor failures or data security threats, it makes difficult for forecasting models to have satisfactory predictions[5-7]. Deletion of abnormal or missing data segments would directly lead to wastage of data[8]. Scholars have already been constructing complete observed space by means of mining the latent structure of sparse data [9-24].

Adaptive Ensemble Learning with Gaussian Copula (GC-AEL) not only incorporates many of the advantages of single models, but also has its own additional innovations, which includes:

- Using the GC to fill up sparsity data, GC-Completion (*GC-Comp.*), the resulting dataset is as realistic as possible and turns it into a serviceable resource.
- Leveraging the principle of adaptive ensemble strategy to integrates the advantages of variant individual models, which sifts out low precision data.

## 2 Related Work

ML and neural networks are gradually being applied to the electric LF field[25]. They are distinguished by their excellent performance in handling depth features with strong nonlinear trends[26]. For example, Chen Chang's fusion of the Prophet model with the XGBOOST model has led to more satisfactory results in LF[27]. The SVM model modified by Sparrow's algorithm (SSA) has a higher accuracy than the traditional model in predicting the electricity price trend[28]. Based on similarity filter (SF), Jialin L. et al. performed forecasting of power usage sequences after fusion optimization of CNN and LSTM (SFCL) [29]. Experiments proved that SFCL has good prediction results. In fact, the intrinsic high correlation characteristics vary across countries and regions, which cannot be fully captured by a single learning model, resulting in limited robustness[30]. Ensemble modeling is an excellent solution path and has been well-practiced. For example, Tinghui Chen combined multiple evolutionary computing algorithms by an ensemble strategy for accurate robot calibration[31]. D. Wu integrated inner produce space and distance space with ensemble algorithm to obtain a highly accurate Web service QoS prediction[32].

Data sparsity [33-38], in practice, will occur. To solve these problems, Wei et al. provides marginal Pearson correlation and marginal rank correlation based on the distribution of missing data[39]. Xin Luo et al. adopts a Single Latent Factor-dependent, Non-negative, Multiplicative and Momentum-incorporated Update algorithm with fast convergence to deal high-dimensional and sparse (HiDS) matrix problems[40]. Di Wu et al. proposed a latent-factor-analysis-based online sparse-streaming-feature selection algorithm to address online streaming feature selection with sparsity[41]. Besides, Di's team also mentioned a latent factor model as an efficient approach to solve HiDS matrix difficulties[42].

## 3. Methodology

### 3.1 Gaussian Copula

*3.1.1 Hypothesis:* It defined that $\mathbf{x}=(x_1,\cdots,x_q)\in\mathbb{R}^q$ is a $q$-dimensional continuous random vector. We denote by $\mathbf{x}_I$ the subvector of $\mathbf{x}$, where $I\subset[\mathcal{M},\mathcal{C},\mathcal{D}]\subset[q]$. Let $\mathcal{M},\mathcal{C},\mathcal{D}$ donate missing, observed continuous, observed ordinal tables. Let $\mathcal{O}=\mathcal{C}\cup\mathcal{D}$ donates observed tables, so $\mathbf{x}=(x_{\mathcal{M}},x_{\mathcal{O}})$. Define $\mathbf{x}^i, X_j, x^i_j$ to denote the whole of the *i*-th row, the whole of the *j*-th column, and the coordinate (*i*, *j*) element.

*3.1.2 Definition of Gaussian Copula:* Suppose that rows of sparse matrix are $a^1, a^2, ..., a^m \sim GC(\Sigma, g)$ and $a^i=(a^i_O, a^i_M)$, $i\in[1,m]$. We initialize the $(\hat{\Sigma}, \hat{g})$ and compute constraints $z^i_{O_i}\in\hat{g}^{-1}_{O_i}(a^i_{O_i}), i=1,...,m$ based on the observed data $\{a^i_{O_i}\}^m_{i=1}$. Then we can solve the missing data $\hat{a}^i_{M_i}$.

*3.1.3 Expectation Maximization (EM) Algorithm:* In E-step, we replace unknown $z^i(z^i)^T$ with expectation conditional on observed value $a^i_{O_i}$ and the $\hat{\Sigma}$. In M-step, we update the estimate of the correlation matrix to the conditional expectation of its covariance. The algorithm is shown in Table 1.

Table 1 EM Algorithm

| 1 | **Input:** Observation $\{a^i_{O_i}\}^m_{i=1}$. |
|---|---|
| 2 | **Repeat** |
| 3 |    **For** $t=1,...,endingTime$ |
| 4 |       $G^{(t)}=G(\Sigma^{(t)}, a^i_{O_i})$   /*E-step*/ |
| 5 |       $\Sigma^{(t+1)}=G^{(t)}$   /*M-step*/ |
| 6 |       $\Sigma^{(t+1)}=P_\Omega(\Sigma^{(t+1)})$   /*Correlation Matrix Scaling*/ |
| 7 |    **End For** |
| 8 | **Until convergence** |
| 9 | **Output:** $\hat{\Sigma}=\Sigma^{(t)}$ |

### 3.2 Basis ML Models

*LSTM:* The main gating components of LSTM[43] are Forget Gate, Input Gate and Output Gate.

$$L_o = \sigma(\omega_o[\hat{D}, h(t-1)]) \tag{1}$$

where $L_o$ is the output of LSTM, $\sigma$ is the activation function.

*CNN:* Convolutional Neural Network (CNN) is a feed-forward neural network[44,45].

$$CN_o = c_P(Con_1(\hat{D}), Con_2(\hat{D})) \tag{2}$$

where $Con(\cdot)$ is the convolution operation and $c_P(\cdot)$ is the pooling function.

*TCN:* Temporal Convolutional Networks (TCN) are a variant structure of CNN[46].

$$F(s)=(X*vf)(s)=\sum_{i=0}^{j-1}f(i)X_{s-vi} \tag{3}$$

where $X$ is the input sequence, $v$ is the dilation convolution operator, and $f$ is the convolution kernel.

*XGBoost:* XGBoost[47] is a GBDT-based optimisation model, which continuous-iteratively optimises sub-trees.

$$F_{XGB}=\min loss(\hat{D}, \bar{D})+\sum_j\Phi(f_j) \tag{4}$$

*TRMF:* In order to make the fitting error between the original and predictions, TRMF improves the accuracy of prediction by making $X\approx\Lambda S$ [48].

$$F_{TR}=\min_{para.} loss(\hat{D}, \Lambda S|_{\hat{D}})+(\sum\lambda\kappa)|_{para.} \tag{5}$$

where *para.* represents the parameters of loss function, $\lambda$ and $\kappa$ represent weight hyperparameters and time series regular phases of $\hat{D}$ with matching *para*.

### 3.3 Adaptive Ensemble

Adaptive ensemble is an effective method for aggregating variant models. Let $err^t(r)$ be the prediction error of the *t*-th model among the prediction models at round $r$, where $t\in N$ and $N$ represents the number of basis models. $CE^t(r)$ be the cumulative error value of this model at step $r$ and before, and $w^t(r)$ be the adaptive weight of this model at round $r$. The mathematical definitions are shown as follows:

$$CE^t(r)=\sum_{l'=1}^r err^t(l') \tag{6}$$

$$w^t(r)=\frac{e^{-\lambda CE^t(r)}}{\sum_{n=1}^N e^{-\lambda CE^t(r)}} \tag{7}$$

where, $\lambda=\sqrt{1/\ln R}$ is the equilibrium factor that controls the aggregation weights, and $R$ is the final round of the *t*-th model from the start of training to convergence. Finally, it weights and sum the forecast values of all models.

$$\tilde{y}_r=\sum_{r=1}^R w^t(r)\hat{y}^t_r \tag{8}$$

where $\hat{y}^t_r$ represents the predicted value of the *r*-th round of the *t*-th model and $\tilde{y}_r$ represents the integrated predicted value. We employ Mean Absolute Percentage Error (Mape) to evaluate the model performance. For further explanations, please refer to the original paper[49].

## 4 Experiments and Results

### 4.1 Experimental Configuration

In this paper, the statistical dataset from an electric power company on the monthly electric load and related features of a region, named D1, is used as the research object. These datasets have 12 features with 01/01/2013 to 31/12/2021 of time range. In order to simulate the data sparsity phenomenon, we randomly erase 10% (marked as D1-0.1) of the data particles. We then make predictions and analyse the results. Next, we showed the important intermediate processes. Finally, the model is subjected to ablation study.

### 4.2 Forecasting Results

We integrate time series models by the Adaptive Ensemble and then compare the results of the ensemble model with those of single models. We rigorously analysed the superiority of the model's predictive results using Mean-Mape, Win/Loss, F-rank, and *p*-value, which are detailed separately under Table 2. It's obvious that the ensemble model is better than the other models at almost all points in time. The ensemble model has the smallest Mean-Mape, and has far more Wins than Losses in all datasets. The ensemble model has the minimum F-rank on all datasets, which means that its accuracy is the highest.

*4.3 Intermediate Process*

In order to visualise the intermediate process, we plot the convergence curves, which are shown in Fig. 1. Due to limitations of spaciousness, only the visualisations for Jan., Apr., Jul. and Oct. are shown. Based on the curves, it can be known that each model eventually converges, and the proposed forecasting model is able to adjust the weights by itself according to the error of each single model and convergence steps.

To demonstrate the effect of sub-modules of the ensemble model, we designed ablation study. In this stage, for every basis-learning model added to the ensemble model sequentially. The order in which the base models are added is dependent on the final forecast error of the single model as well as the adaptive weights. The results are shown in Fig. 2. We found that the overall prediction Mape is progressively smaller with the addition of superior models. However, it sometimes occurs that the Mape remains approximately the same with adding better models. Overall, Adaptive Ensemble is able to self-update the forecasting accuracy by adding superior sub-modules, thus gradually improving the prediction accuracy.

Table 2 Results of Mape on D1-0.1

| Dataset | | D1-0.1 | | | | | |
|---|---|---|---|---|---|---|---|
| Model | | CNN | LSTM | TCN | TRMF | XGB. | Ours |
| Month | Jan. | 6.47% | 2.11% | 5.37% | 7.39% | 14.63% | **2.08%** |
| | Feb. | 38.22% | 5.78% | 2.40% | 13.31% | 36.09% | **2.40%** |
| | Mar. | 11.18% | 11.95% | 14.22% | 3.42% | 9.65% | **3.42%** |
| | Apr. | 7.88% | **1.97%** | 2.21% | 5.97% | 3.05% | 2.10% |
| | May. | 8.10% | 5.09% | 0.76% | 4.57% | 2.08% | **0.76%** |
| | Jun. | 9.80% | 0.32% | 1.43% | 3.32% | 5.03% | **0.32%** |
| | Jul. | 24.50% | 13.77% | 14.27% | 10.37% | **5.51%** | 5.54% |
| | Aug. | 4.18% | 13.93% | 9.40% | 14.94% | 19.16% | **4.18%** |
| | Sept. | 0.33% | 9.93% | 1.69% | 19.66% | 8.18% | **0.33%** |
| | Oct. | 2.68% | 1.93% | 0.87% | 10.68% | 0.67% | **0.67%** |
| | Nov. | 6.61% | 3.53% | 1.93% | 8.38% | 0.01% | **0.01%** |
| | Dec. | 6.09% | 5.30% | 3.90% | **1.66%** | 8.97% | 3.75% |
| Mean-Mape* | | 10.50% ±10.14% | 6.30% ±4.66% | 4.87% ±4.78% | 8.63% ±5.16% | 8.98% ±9.83% | **2.13% ±1.71%** |
| Win/Loss● | | 12/0 | 11/1 | 12/0 | 11/1 | 10/2 | 56/4 |
| F-rank★ | | 4.7500 | 3.6250 | 3.2500 | 4.2083 | 3.5000 | **1.6667** |
| *p*-value♦ | | 0.0010 | 0.0015 | 0.0010 | 0.0010 | 0.0049 | - |

*Mean Mape and their standard deviation; ●Wins/Losses for our model vs. other; ★The F-rank smaller, the higher the accuracy; ♦The p-value test in the Wilcoxon signature rating test with a significance level of 0.05.

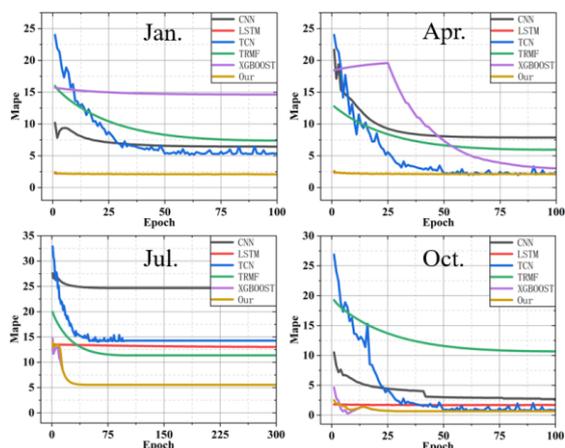

Fig.1 Convergence trends

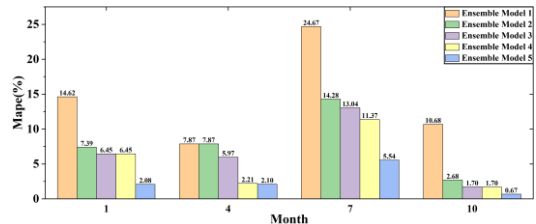

Fig. 2 Results of ablation study on D1-0.1

## 5 Conclusion

This paper proposed a novel hybrid algorithm of Adaptive Ensemble Learning with Gaussian Copula for load forecasting. In practice, even though the sparse dataset complemented by Gaussian Copula enables each ML model to utilise the data and make predictions, the segmentation between different models is too strong to intersect the individual superior path units. Through a large number of experiments, the algorithm proposed in this paper is able to utilize multiple single modules for complementary optimization, which also expands the application scenarios of Gaussian Copula. In the future, we intend to simplify its time complexity.

## 6 Acknowledgement

This work was supported by the Science and Technology Foundation of State Grid Corporation of China under grant 1400-202357341A-1-1-ZN (Identification of Energy Security Risks and Strategic Path Optimization Technology Research under Global Coal-Oil-Gas-Electricity Coupling in China).

## 7 References

[1] Wu, D., Lin, W.: "Efficient Residential Electric Load Forecasting via Transfer Learning and Graph Neural Networks", IEEE Transactions on Smart Grid, 2023, 14, (3), pp. 2423-2431
[2] Echevarria, A., et al.: "Unleashing Sociotechnical Imaginaries to Advance Just and Sustainable Energy Transitions: The Case of Solar Energy in Puerto Rico", IEEE Transactions on Technology and Society, 2023, 4, (3), pp. 255-268
[3] Wu, D., Luo, X., Shang, M., He, Y., Wang, G., and Zhou, M.: "A Deep Latent Factor Model for High-Dimensional and Sparse Matrices in Recommender Systems", IEEE Transactions on Systems, Man, and Cybernetics: Systems, 2021, 51, (7), pp. 4285-4296
[4] Wu, D., Hu, Y., Liu, K., et al.: "An Outlier-Resilient Autoencoder for Representing High-Dimensional and Incomplete Data", IEEE Transactions on Emerging Topics in Computational Intelligence, 2025, 9, (2), pp. 1379-1391
[5] Wu, D., He, Q., Luo, X., Shang, M., He, Y., and Wang, G.: "A Posterior-Neighborhood-Regularized Latent Factor Model for Highly Accurate Web Service QoS Prediction", IEEE Transactions on Services Computing, 2022, 15, (2), pp. 793-805
[6] Yuan, Y., He, Q., Luo, X., and Shang, M.: "A Multilayered-and-Randomized Latent Factor Model for High-Dimensional and Sparse Matrices", IEEE Transactions on Big Data, 2022, 8, (3), pp. 784-794
[7] Luo, X., Wang, Z., and Shang, M.: "An Instance-Frequency-Weighted Regularization Scheme for Non-Negative Latent Factor Analysis on High-Dimensional and Sparse Data", IEEE Transactions on Systems, Man, and Cybernetics: Systems, 2021, 51, (6), pp. 3522-3532
[8] Wu, D., Luo, X., Shang, M., He, Y., Wang, G., and Wu, X.: "A Data-Characteristic-Aware Latent Factor Model for Web Services QoS Prediction", IEEE Transactions on Knowledge and Data Engineering, 2022, 34, (6), pp. 2525-2538
[9] Wu, D., Li, Z., Yu, Z., He, Y., and Luo, X.: "Robust Low-Rank Latent Feature Analysis for Spatiotemporal Signal Recovery", IEEE

Transactions on Neural Networks and Learning Systems, 2025, 36, (2), pp. 2829-2842

[10] Wu, H., Qiao, Y., and Luo, X.: "A Fine-Grained Regularization Scheme for Non-negative Latent Factorization of High-Dimensional and Incomplete Tensors", IEEE Transactions on Services Computing, 2024, 17, (6), pp. 3006-3021

[11] Wu, D., Zhang, P., He, Y., and Luo, X.: "MMLF: Multi-Metric Latent Feature Analysis for High-Dimensional and Incomplete Data", IEEE Transactions on Services Computing, 2024, 17, (2), pp. 575-588

[12] Bi, F., He, T., and Luo, X., "A Fast Nonnegative Autoencoder-Based Approach to Latent Feature Analysis on High-Dimensional and Incomplete Data", IEEE Transactions on Services Computing, 2024, 17, (3), pp. 733-746

[13] Bi, F., He, T., Xie, Y., and Luo, X.: "Two-Stream Graph Convolutional Network-Incorporated Latent Feature Analysis", IEEE Transactions on Services Computing, 2023, 16, (4), pp. 3027-3042

[14] Wu, D., He, Y., and Luo, X.: "A Graph-Incorporated Latent Factor Analysis Model for High-Dimensional and Sparse Data", IEEE Transactions on Emerging Topics in Computing, 2023, 11, (4), pp. 907-917

[15] Chen, J., Wang, R., Wu, D., and Luo, X.: "A Differential Evolution-Enhanced Position-Transitional Approach to Latent Factor Analysis", IEEE Transactions on Emerging Topics in Computational Intelligence, 2023, 7, (2), pp. 389-401

[16] Yuan, Y., Luo, X., Shang, M., and Wang, Z.: "A Kalman-Filter-Incorporated Latent Factor Analysis Model for Temporally Dynamic Sparse Data", IEEE Transactions on Cybernetics, 2023, 53, (9), pp. 5788-5801

[17] Wu, D., Luo, X.: "Robust Latent Factor Analysis for Precise Representation of High-Dimensional and Sparse Data", IEEE/CAA Journal of Automatica Sinica, 2021, 8, (4), pp. 796-805

[18] Luo, X., Zhou, M., Li, S., et al.: "Non-negativity constrained missing data estimation for high-dimensional and sparse matrices from industrial applications", IEEE transactions on cybernetics, 2019, 50, (5), pp. 1844-1855

[19] Song, Y., Li, M., Luo, X., et al.: "Improved symmetric and nonnegative matrix factorization models for undirected, sparse and large-scaled networks: A triple factorization-based approach", IEEE Transactions on Industrial Informatics, 2019, 16, (5), pp. 3006-3017

[20] Shang, M., Luo, X., Liu, Z., et al.: "Randomized latent factor model for high-dimensional and sparse matrices from industrial applications", IEEE/CAA Journal of Automatica Sinica, 2018, 6, (1), pp. 131-141

[21] Luo, X., Zhou, M., Li, S., et al.: "An inherently nonnegative latent factor model for high-dimensional and sparse matrices from industrial applications", IEEE Transactions on Industrial Informatics, 2017, 14, (5), pp. 2011-2022

[22] Luo, X., Sun, J., Wang, Z., et al.: "Symmetric and nonnegative latent factor models for undirected, high-dimensional, and sparse networks in industrial applications", IEEE Transactions on Industrial Informatics, 2017, 13, (6), pp. 3098-3107

[23] Luo, X., Zhou, M., Li, S., et al.: "A nonnegative latent factor model for large-scale sparse matrices in recommender systems via alternating direction method", IEEE transactions on neural networks and learning systems, 2015, 27, (3), pp. 579-592

[24] Luo, X., Zhou, M., Li, S., et al.: "An efficient second-order approach to factorize sparse matrices in recommender systems", IEEE transactions on industrial informatics, 2015, 11, (4), pp. 946-956

[25] Deng, S., Chen, F., Wu, D., He, Y., Ge, H., and Ge, Y.: "Quantitative combination load forecasting model based on forecasting error optimization", Computers and Electrical Engineering, 2022, 101, pp.108125

[26] Wu, D., Sun, W., He, Y., Chen, Z., and Luo, X.: "MKG-FENN: A multimodal knowledge graph fused end-to-end neural network for accurate drug–drug interaction prediction", Proceedings of the AAAI Conference on Artificial Intelligence, 2024, 38, (9), pp. 10216-10224

[27] Chang, C.: "Electricity Consumption Forecasting Based on Improved Prophet Modeling", 2023 IEEE 6th International Conference on Information Systems and Computer Aided Education, China, 2023, pp. 885-889

[28] Duan, Z., Liu, T.: "Short-Term Electricity Price Forecast Based on SSA-SVM Model. Smart Innovation, Systems and Technologies", Advanced Intelligent Technologies for Industry, Singapore, 2022, 285, pp. 79–88

[29] Liu, J., Lu, L., Yu, X. and Wang, X.: "SFCL: Electricity Consumption Forecasting of CNN-LSTM Based on Similar Filter", 2022 China Automation Congress, China, 2022, pp. 4171-4176

[30] Luo, X., Li, Z., Yue, W., and Li, S.: "A Calibrator Fuzzy Ensemble for Highly-Accurate Robot Arm Calibration", IEEE Transactions on Neural Networks and Learning Systems, 2025, 36, (2), pp. 2169-2181

[31] Chen, T., Li, S., Qiao, Y., and Luo, X.: "A Robust and Efficient Ensemble of Diversified Evolutionary Computing Algorithms for Accurate Robot Calibration", IEEE Transactions on Instrumentation and Measurement, 2024, 73, pp. 1-14

[32] Wu, D., Zhang, P., He, Y., and Luo, X.: "A Double-Space and Double-Norm Ensembled Latent Factor Model for Highly Accurate Web Service QoS Prediction", IEEE Transactions on Services Computing, 2023, 16, (2), pp. 802-814

[33] Luo, X., Zhou, M., Xia, Y., Zhu, Q., Ammari, A. C., and Alabdulwahab, A.: "Generating highly accurate predictions for missing QoS data via aggregating nonnegative latent factor models", IEEE Transactions on Neural Networks and Learning Systems, 2015, 27(3), 524-537.

[34] Luo, X., Zhou, M., Xia, Y., and Zhu, Q.: "An efficient non-negative matrix-factorization-based approach to collaborative filtering for recommender systems", IEEE Transactions on Industrial informatics, 2014, 10(2), 1273-1284.

[35] Wu, D., Luo, X., Wang, G., Shang, M., Yuan, Y., and Yan, H.: "A highly accurate framework for self-labeled semisupervised classification in industrial applications", IEEE Transactions on Industrial Informatics, 2017, 14(3), 909-920.

[36] Luo, X., and Zhou, M.: "Effects of extended stochastic gradient descent algorithms on improving latent factor-based recommender systems", IEEE Robotics and Automation Letters, 2019, 4(2), 618-624.

[37] Li, W., He, Q., Luo, X., and Wang, Z.: "Assimilating Second-Order Information for Building Non-Negative Latent Factor Analysis-Based Recommenders", IEEE Transactions on Systems Man Cybernetics: Systems, 2021, 52(1): 485-497.

[38] Luo, X., Liu, Z., Li, S., Shang, M., and Wang, Z.: "A Fast Non-negative Latent Factor Model based on Generalized Momentum Method", IEEE Transactions on Systems Man Cybernetics: Systems, 2021, 51(1): 610-620.

[39] Chen, F., Wu, D., Yang, J., et al.: "An Online Sparse Streaming Feature Selection Algorithm". arXiv preprint arXiv:2208.01562, 2022.

[40] Luo, X., Zhou, Y., Liu, Z., and M. Zhou.: "Fast and Accurate Non-Negative Latent Factor Analysis of High-Dimensional and Sparse Matrices in Recommender Systems", IEEE Transactions on Knowledge and Data Engineering, 2023, 35, (4), pp. 3897-3911

[41] Wu, D., He, Y., Luo, X., and Zhou, M.: "A Latent Factor Analysis-Based Approach to Online Sparse Streaming Feature Selection", IEEE Transactions on Systems, Man, and Cybernetics: Systems, 2022, 52, (11), pp. 6744-6758

[42] Wu, D., Luo, X., He, Y., and Zhou, M.: "A Prediction-Sampling-Based Multilayer-Structured Latent Factor Model for Accurate Representation to High-Dimensional and Sparse Data", IEEE Transactions on Neural Networks and Learning Systems, 2024, 35, (3), pp. 3845-3858

[43] Rubasinghe, O., Zhang, X., Chau, T. K., Chow, Y. H., Fernando, T., and Iu, H.H.-C.: "A Novel Sequence to Sequence Data Modelling Based CNN-LSTM Algorithm for Three Years Ahead Monthly Peak Load Forecasting", IEEE Transactions on Power Systems, 2024, 39, (1), pp. 1932-1947

[44] Wang, H., Li G., Wang, G., Peng, J., Jiang, H., and Liu, Y.: "Deep learning based ensemble approach for probabilistic wind power forecasting", Appl. energy, 2017, 188, pp. 56-70

[45] Ameer, H.K., Shuai, L., and Xin, L.: "Obstacle Avoidance and Tracking Control of Redundant Robotic Manipulator: An RNN based Metaheuristic Approach". IEEE Transactions on Industrial Informatics, 2020, 16(7): 4670-4680.

[46] Luo, D., et al.: "Prediction for Dissolved Gas in Power Transformer Oil Based on TCN and GCN", IEEE Transactions on Industry Applications, 2022, 58, (6), pp. 7818-7826

[47] Chen, T., and Guestrin, C.: "XGBoost: A Scalable Tree Boosting System", Proceedings of the 22nd ACM SIGKDD International Conference on Knowledge Discovery and Data Mining. Association for Computing Machinery, USA, 2016, pp. 785–794

[48] Yu, H.F., Rao, N., and Dhillon, I.S.: "Temporal regularized matrix factorization for high-dimensional time series prediction", Proceedings of the 30th International Conference on Neural Information Processing Systems. Curran Associates Inc., Red Hook, USA, 2016, pp. 847–855

[49] Liang, C., Wu, D., He, Y., Huang, T., Chen, Z., Luo, X.: "MMA: Multi-Metric-Autoencoder for Analyzing High-Dimensional and Incomplete Data", Machine Learning and Knowledge Discovery in Databases: Research Track. 2023, 14173, pp. 3-19